%% file: MICCAI2026-main-conference-paper-template.tex
\documentclass[runningheads]{llncs}
\usepackage[T1]{fontenc}
\usepackage{graphicx,verbatim}
\usepackage{multirow}
\usepackage{amsmath,amssymb}
\begin{document}
%
\title{CLoPA: Continual Low-Parameter Adaptation of Interactive Segmentation for Medical Image Annotation}

%

\author{Parhom Esmaeili\inst{1} \and Chayanin Tangwiriyasakul\inst{1} \and Eli Gibson\inst{2} \and Sebastien Ourselin\inst{1} \and M. Jorge Cardoso\inst{1}}
\authorrunning{P. Esmaeili et al.}
\institute{School of Biomedical Engineering and Imaging Sciences, KCL, UK
\and Siemens Healthineers, Princeton NJ, USA \\
    \email{parhom.esmaeili@kcl.ac.uk}}

\maketitle              
\begin{abstract}


Interactive segmentation enables clinicians to guide annotation, but existing zero-shot models like nnInteractive fail to consistently reach expert-level performance across diverse medical imaging tasks. Because annotation campaigns produce a growing stream of task-specific labelled data, online adaptation of the segmentation model is a natural complement to zero-shot inference. We propose CLoPA, a continual adaptation strategy that tunes a small fraction of nnInteractive's parameters on the annotation cache, triggered by lightweight episode scheduling. CLoPA requires no new parameters or changes to the inference pipeline, and operates entirely within the existing annotation workflow. Across eight Medical Segmentation Decathlon tasks spanning diverse anatomical targets and imaging characteristics, CLoPA rapidly elevates performance to expert-level, even for tasks where nnInteractive previously failed, with the majority of gains realised after a single training episode. We show that the benefits of tuning different parameter groups depends on task characteristics and data regimes. Also, that for targets with complex geometries (e.g., hepatic vessels), instance normalisation and low-level feature tuning saturates, suggesting a need for deeper feature-representation alignment in the most challenging scenarios.

\end{abstract}

\section{Introduction}

Large-scale annotated datasets in medical imaging are bottlenecked by data-sharing restrictions~\cite{deKok2023ARegulation} and the cost of manual segmentation~\cite{Lenchik2019AutomatedReview}. Domain shifts across medical centres further limit the applicability of static pre-trained segmentation models like nnU-Net~\cite{Isensee2021NnU-Net:Segmentation} for annotation.
Interactive segmentation addresses this by letting clinicians guide the process through prompts such as clicks, scribbles, or bounding boxes~\cite{Isensee2025NnInteractive:Segmentation}. Among recent models, nnInteractive~\cite{Isensee2025NnInteractive:Segmentation} achieves strong zero-shot generalisation across diverse anatomies. Yet even nnInteractive cannot consistently reach expert-level performance with low-effort click prompts across all tasks~\cite{Esmaeili2026AEvaluation}, making zero-shot models impractical for large-scale annotation campaigns where both speed and reliability are essential.

Because annotation is inherently repetitive, the annotated samples produced during a campaign constitute a growing task-specific dataset that can be exploited for online model adaptation. This motivates a continual learning approach: rather than relying solely on the zero-shot model, we progressively fine-tune it on the annotation stream to close the gap to expert-level performance. Crucially, adaptation must remain lightweight---both to avoid overfitting on the small, growing cache and to preserve the strong initialisation from pre-training.

In this work, we make the following contributions: (1)~we propose CLoPA, a continual adaptation strategy that fine-tunes only a small subset parameters of a base foundation model (e.g., nnInteractive) during the annotation workflow, requiring no new parameters and no changes to inference; (2)~we demonstrate across eight MSD tasks that this lightweight adaptation rapidly achieves expert-level performance, including on tasks where the base model previously failed; and (3)~we extend the interactive segmentation evaluation protocol of Esmaeili et al.~\cite{Esmaeili2026AEvaluation} with trajectory metrics that capture adaptation dynamics over time.

\begin{figure}[t!]
    \caption{Overview of CLoPA. As annotated samples are produced, the annotation cache grows and periodically triggers training episodes that fine-tune only a small subset of nnInteractive's parameters, progressively improving segmentation quality.}
    \centering
    \includegraphics[width=0.9\linewidth]{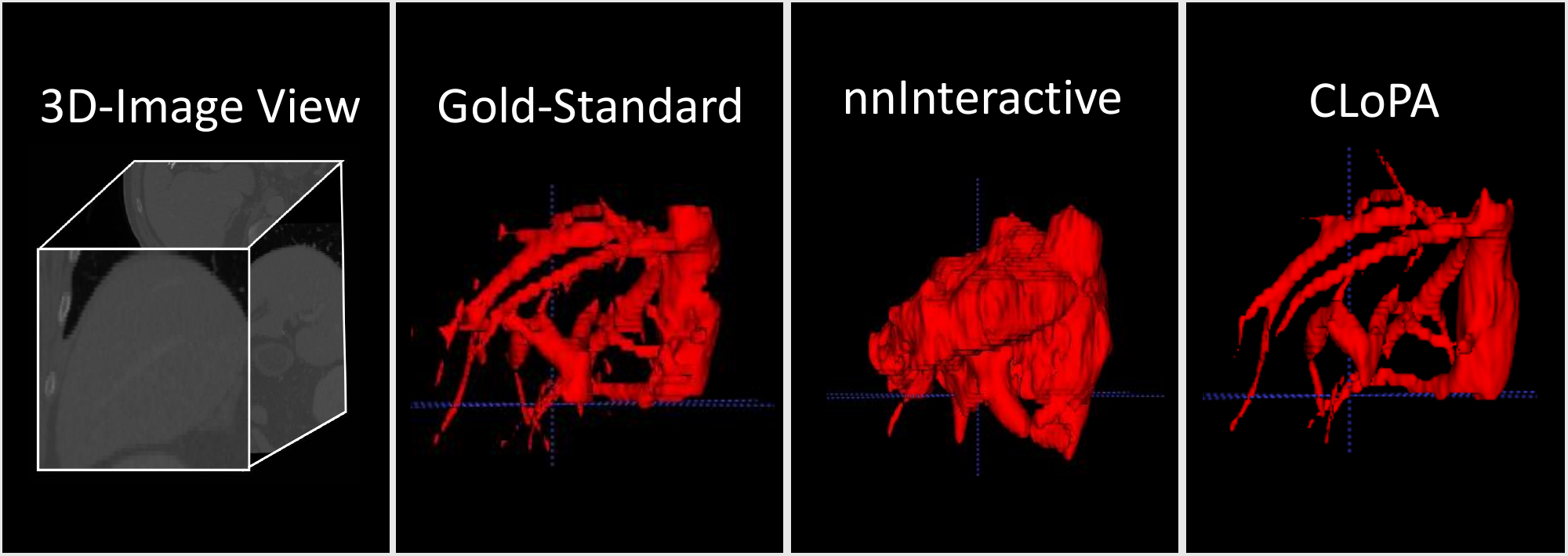}
    \label{fig:graphic_comparison}
\end{figure}

\subsection{Related Work}
Interactive segmentation algorithms incorporate image data and user guided prompts (e.g., clicks, scribbles, bounding boxes or lassos \cite{Isensee2025NnInteractive:Segmentation}) to generate output segmentations on prompted targets. Current state-of-the-art methods employ deep learning to design these algorithms; and during training networks are typically trained with a synthetic user-in-the-loop to iteratively prompt from the error-region between a reference annotation and a prediction \cite{Cheng2023SAM-Med2D,Du2024SegVol:Segmentation,Wang2023SAM-Med3D,Ravi2024SAMVideos,Wong2024ScribblePrompt:Image,Isensee2025NnInteractive:Segmentation}. Zero-shot models, intended for general out-of-the-box use, generate training examples by pooling huge quantities of different segmentation datasets \cite{Wang2023SAM-Med3D,Isensee2025NnInteractive:Segmentation,Cheng2023SAM-Med2D,Du2024SegVol:Segmentation} to sample image-annotation pairs. Training data diversity is introduced by pseudo-labelling strategies: but classical methods like SLIC~\cite{Achanta2012SLICMethods} produce coarse parcellations, while SAM-based projection~\cite{Kirillov2023SegmentAnything,Ravi2024SAMVideos} used by nnInteractive yields mostly blobby instances due to the domain gap from natural images~\cite{Isensee2025NnInteractive:Segmentation}.

Existing methods address prompt ambiguity with multiple candidate masks \cite{Kirillov2023SegmentAnything,Ravi2024SAMVideos}, composite training annotations \cite{Isensee2025NnInteractive:Segmentation}, or more restrictive prompts such as bounding boxes and lassos \cite{Du2024SegVol:Segmentation,Isensee2025NnInteractive:Segmentation}. However, restrictive prompts are laborious for complex geometries, and no current method reliably achieves expert level performance using only low-effort click interactions~\cite{Esmaeili2026AEvaluation}.
Additionally, fixed patch sizes and zooming heuristics~\cite{Du2024SegVol:Segmentation,Isensee2025NnInteractive:Segmentation} used for volumetric inference are suboptimal for targets with low volume fractions (sparse, branching structures), where CNN based methods particularly struggle~\cite{Isensee2025NnInteractive:Segmentation}.

In summary, zero-shot models lack the inductive biases to consistently reach expert-level performance, making them unsuited for large-scale annotation. Since annotation is inherently repetitive, adaptation to the task context is a natural complement. In-context learning approaches~\cite{Wong2024MultiverSeg:Guidance} plateau below specialist performance and do not scale with dataset size. Continual parameter tuning has been explored in 2D~\cite{Xu2025YouImaging}, but to our knowledge has not been extended to volumetric interactive segmentation.

\section{Methodology}

We consider a practical data-annotation scenario: a single, fixed binary segmentation task where annotated samples are cached as they are produced. In this work, we adapt a base interactive-segmentation foundation model, namely nnInteractive~\cite{Isensee2025NnInteractive:Segmentation}, on a stream of incoming annotations using full memory replay, without modifying the inference pipeline. CLoPA has two components: (1)~\emph{when} to trigger adaptation (training episodes), and (2)~\emph{how} to adapt. A training episode is triggered once the annotation cache contains at least 25\% of the dataset ($k_{\mathcal{D}} = 0.25$) and at least 5 unassigned samples (from $k_{\mathcal{M}} = 0.2$, i.e., $\geq 1/k_{\mathcal{M}}$ samples to allow for a validation split).

\subsection{Training Configurations}

\textbf{Tunable parameters:} We freeze all pretrained weights bar two configurations. \textbf{1)} unfreezing only instance normalisation~\cite{Ulyanov2017InstanceStylization} affine parameters (scale and bias); we denote this configuration \textbf{CLoPA-I.N} in all tables. \textbf{2)} tuning all instance normalisation parameters and the convolution kernels in first stage of the U-Net's \cite{Ronneberger2015U-Net:Segmentation} encoder and the last stage of the decoder (segmentation layers); we denote this configuration \textbf{CLoPA-C.N}. Instance normalisation modulates per-channel feature statistics independently for each sample, making it a natural fit for task-specific style and contrast adaptation without altering the learned spatial filters. Because these parameters constitute a tiny fraction of the total model ($<\!0.01\%$), the risk of catastrophic forgetting or overfitting on the small annotation cache is minimal, while still providing capacity to recalibrate feature distributions for the target anatomy. Convolution kernels determine the learned feature representations; making it a natural fit for more challenging target anatomies. Tuning at shallow depths allows for aligning low-level features while avoiding overfitting.

\noindent \textbf{Data synthesis:} Following nnU-Net~\cite{Isensee2021NnU-Net:Segmentation}, we sample $192^3$ patches uniformly across foreground and background classes from the annotation cache, applying nnInteractive's preprocessing and augmentation pipeline. For each gradient step, we simulate click-based interaction: one foreground and one background point sampled from false-negative regions per interaction step. 

We currently employ fine-tuning with only click-prompting as this is the least laborious form of user-interaction. For training we use an unweighted Dice Cross-Entropy loss \cite{Jadon2020ASegmentation} averaged across interaction time-steps, same as Wong et al. \cite{Wong2024ScribblePrompt:Image}. For a maximum number of interaction time-steps, $N$ our loss is:
\begin{equation}
    L = \frac{1}{N}\sum_{i=1}^{N} L_{Dice}(m_{i}, y) + L_{CE}(m_{i}, y)
\label{eq:loss}
\end{equation}
Where $m_{i}$ indicates the softmaxed prediction for the batch at interaction-step $i$. Each loss term is averaged across batch samples, if a batch sample reaches a termination condition (Dice Score of 1) then it is not accounted for on subsequent iteration time-steps in the loss calculation. For each gradient update we employ $N=5$ interaction steps per gradient update and a fixed initial batch size of 2.  For each training episode, an initial learning rate of $1e^{-3}$ is used for all parameters being fine-tuned with an ADAM optimiser. We train with a fixed quantity of $10$ epochs, and for each epoch, we perform a fixed quantity of $50$ gradient updates.

\section{Experiments}

\textbf{Tasks:} We extend the four axes of task complexity examined by Esmaeili et al. \cite{Esmaeili2026AEvaluation} to the following (conducting evaluations in the native image spaces): \textbf{(i)} Variation in image voxel count (i.e. volume size), \textbf{(ii)} image spacing/anisotropy, \textbf{(iii)} target geometry (spherical vs irregular vs jagged targets occupying low-volume fractions), \textbf{(iv)} target size \textbf{(v)} target detectability and \textbf{(vi)} training dataset size. In line with Esmaeili et al.~\cite{Esmaeili2026AEvaluation} we chose binary semantic segmentation tasks from the Medical Segmentation Decathlon (MSD)~\cite{Antonelli2022TheDecathlon}. For multi-sequence datasets, the sequence that best visualises the target was used. 
\begin{table}[t]
\centering
\caption{Task characteristics. All tasks are binary semantic segmentation from the MSD~\cite{Antonelli2022TheDecathlon}. Dataset sizes are post-split.}
\resizebox{0.9\linewidth}{!}{
\begin{tabular}{lclr}
\hline
Task & Sequence & Key characteristics & $N$ \\
\hline
Hippocampus & T1 & Small volume, fine detail & 130 \\
Brain tumour core & T2w & Irregular, ambiguous boundaries & 240 \\
Pancreas & CT & Large, blobby & 140 \\
Liver & CT & Very large, blobby, easy detection & 65 \\
Prostate & T2w & Spherical, highly anisotropic & 16 \\
Lung lesion & CT & Small target in large volume & 31 \\
Hepatic vessels & CT & Sparse, branching, low volume fraction & 151 \\
Colon cancer & CT & Hard to detect & 63 \\
\hline
\end{tabular}}
\label{tab:tasks}
\end{table}

\noindent\textbf{Prompting:} As with Esmaeili et al. \cite{Esmaeili2026AEvaluation}, all simulations use one point per iteration per class randomly sampled from false-negative foreground and background regions. Simulations consist of interactive initialisation and 100 editing steps. 

\noindent\textbf{Model Evaluation:} We extend the evaluation metrics of Esmaeili et al.~\cite{Esmaeili2026AEvaluation}. On a per-sample basis, we compute Dice and Normalised Surface Dice (NSD) with MSD~\cite{Antonelli2022TheDecathlon} tolerances at each interaction step, as well as interaction-count-normalised AUCs for both metrics (nAUC). Following Esmaeili et al.~\cite{Esmaeili2026AEvaluation}, we estimate the number of interactions to expert-level performance (NoI) on a per-sample Dice basis, normalised by the maximum interaction count to obtain a percentage (nNoI). Expert-level thresholds are defined as the task-specific mean Dice of nnU-Net~\cite{Isensee2021NnU-Net:Segmentation} trained on the full training set with optimal configuration selection. All expected metric values are reported as dataset-wide means. The percentage of samples that did not reach the performance target (NoF) is also reported. For adaptation studies, we extend the evaluation by tracking the trajectory of each metric's expected performance $\mathbb{E}[m_{j}(t)]$ as a function of dataset size $t$, evaluating after each training episode. We report AUCs over these trajectories (trajectory AUCs). Significance rankings follow the MSD protocol~\cite{Antonelli2022TheDecathlon} with pairwise algorithm comparisons: Wilcoxon signed-rank tests for all metrics, except episodic NoF where a McNemar test is used since comparisons are between binary outcomes. For trajectory measures, pairing is performed along data-sample index rather than test samples. Significance is reported at the $\alpha=0.05$ level.

\noindent\textbf{Data Splitting and Evaluation Runs:} MSD training datasets were split 50-50 into a train and holdout set. For each training run the adaptive algorithms iterate through the training set; with the sequence of the training data stream permuted across 3 runs, to obtain 3 runs of episodic checkpoints. Inference is also performed on 3 runs to simulate prompting stochasticity, with the corresponding training-run model checkpoints. For static models we use the same checkpoint across inference runs. For episodic performance measures we first average metrics across runs before reporting statistics. For trajectory measures, we obtain the trajectories of expected performances and then average across training runs, as training is not generally triggered synchronously across runs.


\section{Results and Discussion}\label{'Results and Discussion'}

\input{results_tex_files/main_results}



\bibliographystyle{splncs04}
\bibliography{references}
\end{document}

%% file: results_tex_files/main_results.tex
\begin{table}[h!]
\caption{Final-episode performance. CLoPA-I.N: instance normalisation (I.N) only, CLoPA-C.N: Instance-normalisation and shallow depth convolution kernel tuning. All metrics report means except NoF (percentage of failed samples). Bold indicates first-ranked method per metric and task according to significance rankings.}
\centering
\resizebox{0.915\textwidth}{!}{
\begin{tabular}{|c|c|ccc|ccc|cc|}
\hline
\multirow{2}{*}{Task} 
 & \multirow{2}{*}{Algorithm} 
 & \multicolumn{3}{c|}{Dice} 
 & \multicolumn{3}{c|}{NSD} 
 & \multicolumn{2}{c|}{NoI} \\ \cline{3-10}
 
 &  
 & \multicolumn{1}{c|}{ Init.} 
 & \multicolumn{1}{c|}{ Iter. 100} 
 & nAUC 
 & \multicolumn{1}{c|}{ Init.} 
 & \multicolumn{1}{c|}{ Iter. 100} 
 & nAUC 
 & \multicolumn{1}{c|}{nNoI} 
 & NoF \\ \hline
\multirow{3}{*}{Brain Tumour}  
 &  nnInteractive  
 & \multicolumn{1}{c|}{ 0.505 } 
 & \multicolumn{1}{c|}{ 0.753 } 
 &  0.742  
 & \multicolumn{1}{c|}{ 0.583 } 
 & \multicolumn{1}{c|}{ 0.922 } 
 &  0.904  
 & \multicolumn{1}{c|}{ 29.6 } 
 &  24.4 \\\cline{2-10}
  
 &  CLoPA-I.N  
 & \multicolumn{1}{c|}{ 0.682 } 
 & \multicolumn{1}{c|}{ \textbf{0.815} } 
 &  \textbf{0.817}  
 & \multicolumn{1}{c|}{ 0.812 } 
 & \multicolumn{1}{c|}{ 0.968 } 
 &  0.963  
 & \multicolumn{1}{c|}{ 14.8 } 
 &  12.4 \\\cline{2-10}
  
 &  CLoPA-C.N  
 & \multicolumn{1}{c|}{ \textbf{0.694} } 
 & \multicolumn{1}{c|}{ 0.811 } 
 &  0.814  
 & \multicolumn{1}{c|}{ \textbf{0.827} } 
 & \multicolumn{1}{c|}{ 0.968 } 
 &  \textbf{0.964}  
 & \multicolumn{1}{c|}{ 15.4 } 
 &  13.2 \\\hline
\multirow{3}{*}{Liver}  
 &  nnInteractive  
 & \multicolumn{1}{c|}{ 0.373 } 
 & \multicolumn{1}{c|}{ 0.970 } 
 &  0.963  
 & \multicolumn{1}{c|}{ 0.384 } 
 & \multicolumn{1}{c|}{ 0.992 } 
 &  0.984  
 & \multicolumn{1}{c|}{ 22 } 
 &  18.2 \\\cline{2-10}
  
 &  CLoPA-I.N  
 & \multicolumn{1}{c|}{ \textbf{0.912} } 
 & \multicolumn{1}{c|}{ \textbf{0.974} } 
 &  \textbf{0.973}  
 & \multicolumn{1}{c|}{ \textbf{0.910} } 
 & \multicolumn{1}{c|}{ \textbf{0.994} } 
 &  \textbf{0.992}  
 & \multicolumn{1}{c|}{ \textbf{11.7} } 
 &  \textbf{6.06} \\\cline{2-10}
  
 &  CLoPA-C.N  
 & \multicolumn{1}{c|}{ 0.848 } 
 & \multicolumn{1}{c|}{ 0.971 } 
 &  0.969  
 & \multicolumn{1}{c|}{ 0.848 } 
 & \multicolumn{1}{c|}{ 0.991 } 
 &  0.988  
 & \multicolumn{1}{c|}{ 15.9 } 
 &  10.6 \\\hline
\multirow{3}{*}{Hippocampus}  
 &  nnInteractive  
 & \multicolumn{1}{c|}{ 0.585 } 
 & \multicolumn{1}{c|}{ 0.805 } 
 &  0.788  
 & \multicolumn{1}{c|}{ 0.631 } 
 & \multicolumn{1}{c|}{ 0.890 } 
 &  0.870  
 & \multicolumn{1}{c|}{ 99 } 
 &  97.7 \\\cline{2-10}
  
 &  CLoPA-I.N  
 & \multicolumn{1}{c|}{ 0.875 } 
 & \multicolumn{1}{c|}{ 0.903 } 
 &  0.902  
 & \multicolumn{1}{c|}{ 0.957 } 
 & \multicolumn{1}{c|}{ 0.984 } 
 &  0.983  
 & \multicolumn{1}{c|}{ 45.6 } 
 &  40 \\\cline{2-10}
  
 &  CLoPA-C.N  
 & \multicolumn{1}{c|}{ \textbf{0.876} } 
 & \multicolumn{1}{c|}{ \textbf{0.911} } 
 &  \textbf{0.909}  
 & \multicolumn{1}{c|}{ 0.957 } 
 & \multicolumn{1}{c|}{ \textbf{0.988} } 
 &  \textbf{0.987}  
 & \multicolumn{1}{c|}{ \textbf{38} } 
 &  \textbf{31.5} \\\hline
\multirow{3}{*}{Prostate}  
 &  nnInteractive  
 & \multicolumn{1}{c|}{ 0.774 } 
 & \multicolumn{1}{c|}{ 0.922 } 
 &  0.921  
 & \multicolumn{1}{c|}{ 0.815 } 
 & \multicolumn{1}{c|}{ 0.985 } 
 &  0.983  
 & \multicolumn{1}{c|}{ 4.6 } 
 &  0 \\\cline{2-10}
  
 &  CLoPA-I.N  
 & \multicolumn{1}{c|}{ 0.882 } 
 & \multicolumn{1}{c|}{ 0.935 } 
 &  0.932  
 & \multicolumn{1}{c|}{ 0.941 } 
 & \multicolumn{1}{c|}{ 0.992 } 
 &  0.990  
 & \multicolumn{1}{c|}{ 1.8 } 
 &  0 \\\cline{2-10}
  
 &  CLoPA-C.N  
 & \multicolumn{1}{c|}{ 0.887 } 
 & \multicolumn{1}{c|}{ 0.936 } 
 &  0.932  
 & \multicolumn{1}{c|}{ 0.945 } 
 & \multicolumn{1}{c|}{ \textbf{0.994} } 
 &  0.991  
 & \multicolumn{1}{c|}{ 2 } 
 &  0 \\\hline
\multirow{3}{*}{Lung Lesion}  
 &  nnInteractive  
 & \multicolumn{1}{c|}{ 0.698 } 
 & \multicolumn{1}{c|}{ 0.849 } 
 &  0.852  
 & \multicolumn{1}{c|}{ 0.743 } 
 & \multicolumn{1}{c|}{ 0.940 } 
 &  0.936  
 & \multicolumn{1}{c|}{ 2 } 
 &  0 \\\cline{2-10}
  
 &  CLoPA-I.N  
 & \multicolumn{1}{c|}{ 0.764 } 
 & \multicolumn{1}{c|}{ \textbf{0.861} } 
 &  0.856  
 & \multicolumn{1}{c|}{ 0.819 } 
 & \multicolumn{1}{c|}{ 0.950 } 
 &  0.942  
 & \multicolumn{1}{c|}{ 1.5 } 
 &  0 \\\cline{2-10}
  
 &  CLoPA-C.N  
 & \multicolumn{1}{c|}{ 0.762 } 
 & \multicolumn{1}{c|}{ 0.858 } 
 &  0.855  
 & \multicolumn{1}{c|}{ 0.820 } 
 & \multicolumn{1}{c|}{ 0.946 } 
 &  0.941  
 & \multicolumn{1}{c|}{ 1.6 } 
 &  0 \\\hline
\multirow{3}{*}{Pancreas}  
 &  nnInteractive  
 & \multicolumn{1}{c|}{ 0.454 } 
 & \multicolumn{1}{c|}{ 0.895 } 
 &  0.881  
 & \multicolumn{1}{c|}{ 0.538 } 
 & \multicolumn{1}{c|}{ 0.983 } 
 &  0.971  
 & \multicolumn{1}{c|}{ 9.8 } 
 &  2.13 \\\cline{2-10}
  
 &  CLoPA-I.N  
 & \multicolumn{1}{c|}{ 0.668 } 
 & \multicolumn{1}{c|}{ \textbf{0.897} } 
 &  \textbf{0.888}  
 & \multicolumn{1}{c|}{ 0.760 } 
 & \multicolumn{1}{c|}{ \textbf{0.985} } 
 &  0.979  
 & \multicolumn{1}{c|}{ 8.3 } 
 &  2.84 \\\cline{2-10}
  
 &  CLoPA-C.N  
 & \multicolumn{1}{c|}{ \textbf{0.687} } 
 & \multicolumn{1}{c|}{ 0.895 } 
 &  0.887  
 & \multicolumn{1}{c|}{ \textbf{0.781} } 
 & \multicolumn{1}{c|}{ 0.984 } 
 &  0.978  
 & \multicolumn{1}{c|}{ 8.5 } 
 &  2.84 \\\hline
\multirow{3}{*}{Hepatic Vessels}  
 &  nnInteractive  
 & \multicolumn{1}{c|}{ 0.113 } 
 & \multicolumn{1}{c|}{ 0.165 } 
 &  0.209  
 & \multicolumn{1}{c|}{ 0.122 } 
 & \multicolumn{1}{c|}{ 0.232 } 
 &  0.289  
 & \multicolumn{1}{c|}{ 84.7 } 
 &  82.9 \\\cline{2-10}
  
 &  CLoPA-I.N  
 & \multicolumn{1}{c|}{ 0.511 } 
 & \multicolumn{1}{c|}{ \textbf{0.698} } 
 &  \textbf{0.687}  
 & \multicolumn{1}{c|}{ 0.682 } 
 & \multicolumn{1}{c|}{ \textbf{0.871} } 
 &  \textbf{0.856}  
 & \multicolumn{1}{c|}{ \textbf{19.6} } 
 &  11.8 \\\cline{2-10}
  
 &  CLoPA-C.N  
 & \multicolumn{1}{c|}{ \textbf{0.521} } 
 & \multicolumn{1}{c|}{ 0.692 } 
 &  0.681  
 & \multicolumn{1}{c|}{ \textbf{0.693} } 
 & \multicolumn{1}{c|}{ 0.864 } 
 &  0.849  
 & \multicolumn{1}{c|}{ 23 } 
 &  14.5 \\\hline
\multirow{3}{*}{Colon Cancer}  
 &  nnInteractive  
 & \multicolumn{1}{c|}{ 0.475 } 
 & \multicolumn{1}{c|}{ 0.728 } 
 &  0.725  
 & \multicolumn{1}{c|}{ 0.520 } 
 & \multicolumn{1}{c|}{ 0.788 } 
 &  0.790  
 & \multicolumn{1}{c|}{ 2.4 } 
 &  0 \\\cline{2-10}
  
 &  CLoPA-I.N  
 & \multicolumn{1}{c|}{ 0.626 } 
 & \multicolumn{1}{c|}{ \textbf{0.819} } 
 &  \textbf{0.813}  
 & \multicolumn{1}{c|}{ 0.721 } 
 & \multicolumn{1}{c|}{ \textbf{0.913} } 
 &  \textbf{0.911}  
 & \multicolumn{1}{c|}{ \textbf{1.2} } 
 &  0 \\\cline{2-10}
  
 &  CLoPA-C.N  
 & \multicolumn{1}{c|}{ 0.615 } 
 & \multicolumn{1}{c|}{ 0.781 } 
 &  0.779  
 & \multicolumn{1}{c|}{ 0.716 } 
 & \multicolumn{1}{c|}{ 0.874 } 
 &  0.874  
 & \multicolumn{1}{c|}{ 2 } 
 &  0 \\\hline
\end{tabular}
}\label{table:experiment_main}
\end{table}

Table~\ref{table:experiment_main} compares nnInteractive, CLoPA-I.N and CLoPa-C.N after the final training episode (approximately nnU-Net-equivalent quantity of training data).


\noindent \textbf{Tasks where the base model converges:} For tasks where nnInteractive already achieves consistent convergence to nnU-Net-level performance (NoF~$\lesssim 5\%$: liver, prostate, pancreas, lung lesion, colon cancer)---typically blobby targets or those where detectability is the main challenge---CLoPA-I.N and CLoPA C.N maintains convergence rates (except pancreas, though not at a statistically significant rate) while improving all other metrics. Initialisation Dice and NSD are substantially higher, indicating that task-alignment accelerates annotation efficiency, especially for large targets like liver where adaptation provides sufficient context to trigger nnInteractive's auto-zoom mechanism with minimal prompting. Final-iteration Dice and NSD, as well as nAUC metrics, also improve---particularly for tasks where zero-shot performance was not already saturated, such as colon cancer. Thus, task-alignment raises the performance ceiling and improves editing stability even for easier tasks. For these tasks CLoPA-I.N generally outperforms CLoPA-C.N. This is likely because the existing low-level features are sufficient, and with relatively fewer samples (all these tasks are in medium dataset size at most) instance normalisation tuning is more robust.

\begin{figure}[t]
    \caption{Comparison across methods for the trajectory of mean Dice Score after the editing terminates, as a function of data samples received. Left to right, trajectory on tasks: hepatic vessel, brain tumour core, hippocampus. Vertical lines indicate the number of samples (NoS) to reach nnU-Net performance (horizontal red line).}
    \centering
    \includegraphics[width=0.925\linewidth]{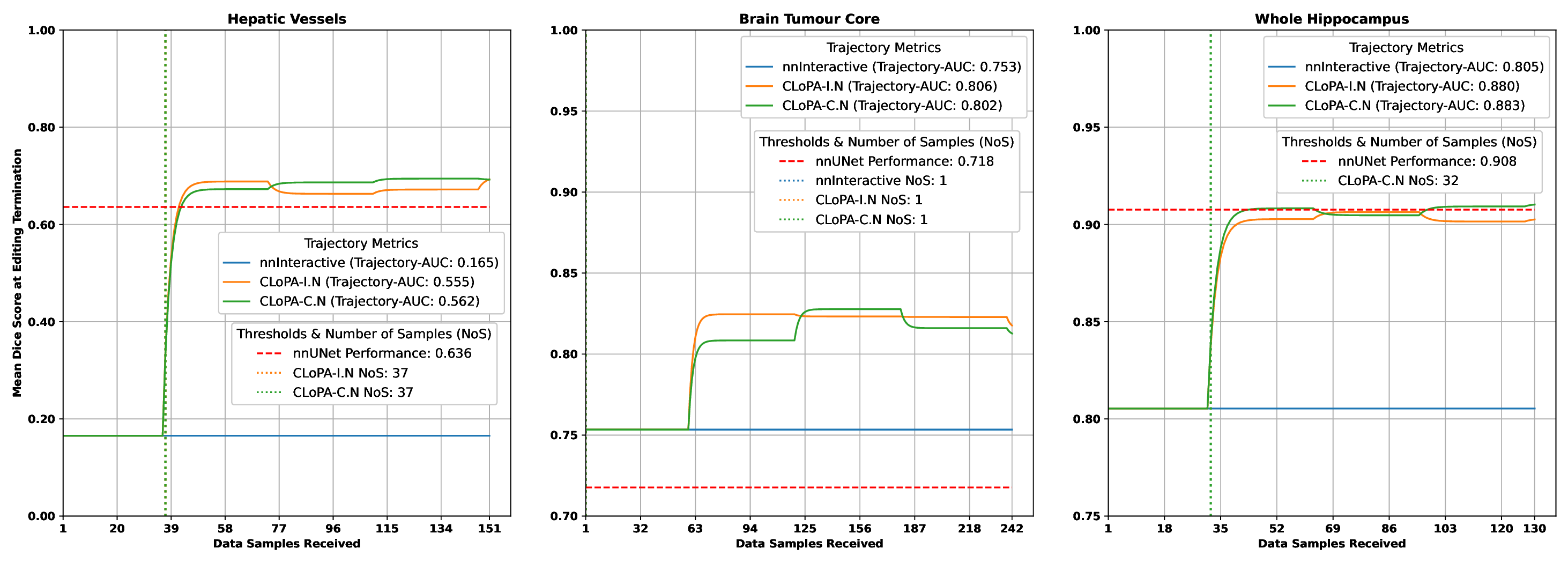}
    \label{fig:final_dice_trajectory_selection}
\end{figure}

\noindent \textbf{Tasks where the base model struggles:} For more challenging tasks (NoF~$\gtrsim 20\%$: brain tumour core, hippocampus, hepatic vessels), adaptation yields large performance gains across all metrics: substantially better initialisation, faster annotation (nNoI), more robust editing stability (nAUC), and higher performance ceilings. These tasks share characteristics that expose limitations in nnInteractive's zero-shot capabilities: ambiguous segmentation boundaries (brain tumour core), fine detail requirements at image sizes deviating from the model's $192^3$ input patch (brain tumour core, hippocampus), and sparse branching structures with low volume fractions (hepatic vessels). For brain tumour core and hippocampus, the base model plateaus early but remains stable (nAUC slightly below final-iteration performance), indicating a performance ceiling when segmenting ambiguous targets with sparse point prompts. Instance normalisation adaptation partially mitigates this but improvements plateau after the first training episode (Fig.~\ref{fig:final_dice_trajectory_selection}), 
likely because the underlying feature representations are not adjusted for ambiguous targets or images smaller than the patch size. Further shallow-depth convolutional kernel tuning noticeably boosts performance for hippocampus, but not for brain tumour core. This is due to an underlying challenge with hippocampus being the small voxel-count, allowing low-level feature tuning to elevate the base model to nnU-Net performance (Fig.~\ref{fig:final_dice_trajectory_selection}). For brain tumour core, ambiguous tissue boundaries likely require deep representation tuning (hence mixed-results with instance normalisation only tuning).

Hepatic vessels present a different challenge: sparse branching structures with low volume fractions, likely underrepresented in nnInteractive's training data. The base model exhibits volatile peak-and-dip editing behaviour (nAUC much higher than final-iteration metrics). Post adaptation, initialisation performance immediately spikes and the editing performance does not exhibit this behaviour. After the first training episode (Fig.~\ref{fig:final_dice_trajectory_selection}), instance normalisation tuning allows expected post-editing performance to exceed nnU-Net performance within approximately 20\% of the editing budget. However, with NoF at $11.8\%$, performance still saturates---as illustrated by the plateauing after the initial episode in Fig.~\ref{fig:final_dice_trajectory_selection}. This saturation likely reflects the inability of instance normalisation alone to learn task-specific feature representations. Contributing factors may include the CNN architecture's limitations for targets with long-range dependencies (requiring adjusting convolution kernels). However, CLoPA-C.N does not provide improvements, indicating that tuning at shallow depth is limited or that the short interaction window during training ($N{=}5$ edits), may be limiting exposure to gradient updates near segmentation completion.

\begin{table}[h]
\caption{Trajectory AUC performance across all tasks. CLoPA-I.N: instance normalisation (I.N) only, CLoPA-C.N: I.N and shallow depth convolution kernel tuning. Metrics summarise AUCs over the trajectory of expected performance as a function of dataset size, except NoF (percentage of failed samples). Bold indicates first-ranked method per metric and task according to significance rankings.}
\centering
\resizebox{0.915\textwidth}{!}{
\begin{tabular}{|c|c|ccc|ccc|cc|}
\hline
\multirow{2}{*}{Task} 
 & \multirow{2}{*}{Algorithm} 
 & \multicolumn{3}{c|}{Dice} 
 & \multicolumn{3}{c|}{NSD} 
 & \multicolumn{2}{c|}{NoI} \\ \cline{3-10}
 
 &  
 & \multicolumn{1}{c|}{ Init.} 
 & \multicolumn{1}{c|}{ Iter. 100} 
 & nAUC 
 & \multicolumn{1}{c|}{ Init.} 
 & \multicolumn{1}{c|}{ Iter. 100} 
 & nAUC 
 & \multicolumn{1}{c|}{nNoI} 
 & NoF \\ \hline
\multirow{3}{*}{Brain Tumour}  
 &  nnInteractive  
 & \multicolumn{1}{c|}{ 0.505 } 
 & \multicolumn{1}{c|}{ 0.753 } 
 &  0.742  
 & \multicolumn{1}{c|}{ 0.583 } 
 & \multicolumn{1}{c|}{ 0.922 } 
 &  0.904  
 & \multicolumn{1}{c|}{ 26.8 } 
 &  20.4 \\\cline{2-10}
  
 &  CLoPA-I.N  
 & \multicolumn{1}{c|}{ 0.623 } 
 & \multicolumn{1}{c|}{ \textbf{0.806} } 
 &  \textbf{0.802}  
 & \multicolumn{1}{c|}{ 0.740 } 
 & \multicolumn{1}{c|}{ \textbf{0.960} } 
 &  0.950  
 & \multicolumn{1}{c|}{ \textbf{15.8} } 
 &  \textbf{11.7} \\\cline{2-10}
  
 &  CLoPA-C.N  
 & \multicolumn{1}{c|}{ \textbf{0.632} } 
 & \multicolumn{1}{c|}{ 0.802 } 
 &  0.799  
 & \multicolumn{1}{c|}{ \textbf{0.751} } 
 & \multicolumn{1}{c|}{ 0.958 } 
 &  0.950  
 & \multicolumn{1}{c|}{ 16.8 } 
 &  13.1 \\\hline
\multirow{3}{*}{Liver}  
 &  nnInteractive  
 & \multicolumn{1}{c|}{ 0.373 } 
 & \multicolumn{1}{c|}{ 0.970 } 
 &  0.963  
 & \multicolumn{1}{c|}{ 0.384 } 
 & \multicolumn{1}{c|}{ 0.992 } 
 &  0.984  
 & \multicolumn{1}{c|}{ 17.5 } 
 &  13.6 \\\cline{2-10}
  
 &  CLoPA-I.N  
 & \multicolumn{1}{c|}{ \textbf{0.734} } 
 & \multicolumn{1}{c|}{ 0.972 } 
 &  \textbf{0.968}  
 & \multicolumn{1}{c|}{ \textbf{0.737} } 
 & \multicolumn{1}{c|}{ 0.992 } 
 &  \textbf{0.988}  
 & \multicolumn{1}{c|}{ \textbf{15.8} } 
 &  10.2 \\\cline{2-10}
  
 &  CLoPA-C.N  
 & \multicolumn{1}{c|}{ 0.717 } 
 & \multicolumn{1}{c|}{ 0.971 } 
 &  0.967  
 & \multicolumn{1}{c|}{ 0.719 } 
 & \multicolumn{1}{c|}{ 0.993 } 
 &  0.987  
 & \multicolumn{1}{c|}{ 16 } 
 &  10 \\\hline
\multirow{3}{*}{Hippocampus}  
 &  nnInteractive  
 & \multicolumn{1}{c|}{ 0.585 } 
 & \multicolumn{1}{c|}{ 0.805 } 
 &  0.788  
 & \multicolumn{1}{c|}{ 0.631 } 
 & \multicolumn{1}{c|}{ 0.890 } 
 &  0.870  
 & \multicolumn{1}{c|}{ 98 } 
 &  94.9 \\\cline{2-10}
  
 &  CLoPA-I.N  
 & \multicolumn{1}{c|}{ 0.803 } 
 & \multicolumn{1}{c|}{ 0.880 } 
 &  0.875  
 & \multicolumn{1}{c|}{ 0.874 } 
 & \multicolumn{1}{c|}{ 0.962 } 
 &  0.956  
 & \multicolumn{1}{c|}{ 55.4 } 
 &  48.6 \\\cline{2-10}
  
 &  CLoPA-C.N  
 & \multicolumn{1}{c|}{ 0.803 } 
 & \multicolumn{1}{c|}{ \textbf{0.883} } 
 &  \textbf{0.878}  
 & \multicolumn{1}{c|}{ 0.874 } 
 & \multicolumn{1}{c|}{ \textbf{0.963} } 
 &  \textbf{0.957}  
 & \multicolumn{1}{c|}{ \textbf{52.7} } 
 &  \textbf{45.8} \\\hline
\multirow{3}{*}{Prostate}  
 &  nnInteractive  
 & \multicolumn{1}{c|}{ 0.774 } 
 & \multicolumn{1}{c|}{ 0.922 } 
 &  0.921  
 & \multicolumn{1}{c|}{ 0.815 } 
 & \multicolumn{1}{c|}{ 0.985 } 
 &  0.983  
 & \multicolumn{1}{c|}{ 4.2 } 
 &  0 \\\cline{2-10}
  
 &  CLoPA-I.N  
 & \multicolumn{1}{c|}{ 0.850 } 
 & \multicolumn{1}{c|}{ 0.932 } 
 &  0.929  
 & \multicolumn{1}{c|}{ 0.899 } 
 & \multicolumn{1}{c|}{ 0.991 } 
 &  0.988  
 & \multicolumn{1}{c|}{ 3.7 } 
 &  0.694 \\\cline{2-10}
  
 &  CLoPA-C.N  
 & \multicolumn{1}{c|}{ 0.851 } 
 & \multicolumn{1}{c|}{ 0.932 } 
 &  \textbf{0.930}  
 & \multicolumn{1}{c|}{ 0.901 } 
 & \multicolumn{1}{c|}{ 0.991 } 
 &  \textbf{0.988}  
 & \multicolumn{1}{c|}{ \textbf{2.8} } 
 &  0 \\\hline
\multirow{3}{*}{Lung Lesion}  
 &  nnInteractive  
 & \multicolumn{1}{c|}{ 0.698 } 
 & \multicolumn{1}{c|}{ 0.849 } 
 &  0.852  
 & \multicolumn{1}{c|}{ 0.743 } 
 & \multicolumn{1}{c|}{ 0.940 } 
 &  0.936  
 & \multicolumn{1}{c|}{ 1.8 } 
 &  0 \\\cline{2-10}
  
 &  CLoPA-I.N  
 & \multicolumn{1}{c|}{ \textbf{0.748} } 
 & \multicolumn{1}{c|}{ \textbf{0.860} } 
 &  \textbf{0.857}  
 & \multicolumn{1}{c|}{ \textbf{0.797} } 
 & \multicolumn{1}{c|}{ \textbf{0.947} } 
 &  \textbf{0.939}  
 & \multicolumn{1}{c|}{ \textbf{1.7} } 
 &  0 \\\cline{2-10}
  
 &  CLoPA-C.N  
 & \multicolumn{1}{c|}{ 0.744 } 
 & \multicolumn{1}{c|}{ 0.857 } 
 &  0.854  
 & \multicolumn{1}{c|}{ 0.795 } 
 & \multicolumn{1}{c|}{ 0.939 } 
 &  0.931  
 & \multicolumn{1}{c|}{ 2 } 
 &  0.243 \\\hline
\multirow{3}{*}{Pancreas}  
 &  nnInteractive  
 & \multicolumn{1}{c|}{ 0.454 } 
 & \multicolumn{1}{c|}{ 0.895 } 
 &  0.881  
 & \multicolumn{1}{c|}{ 0.538 } 
 & \multicolumn{1}{c|}{ 0.983 } 
 &  0.971  
 & \multicolumn{1}{c|}{ 9.3 } 
 &  2.13 \\\cline{2-10}
  
 &  CLoPA-I.N  
 & \multicolumn{1}{c|}{ 0.593 } 
 & \multicolumn{1}{c|}{ \textbf{0.899} } 
 &  \textbf{0.887}  
 & \multicolumn{1}{c|}{ 0.683 } 
 & \multicolumn{1}{c|}{ \textbf{0.986} } 
 &  \textbf{0.977}  
 & \multicolumn{1}{c|}{ \textbf{8.1} } 
 &  \textbf{2.01} \\\cline{2-10}
  
 &  CLoPA-C.N  
 & \multicolumn{1}{c|}{ \textbf{0.600} } 
 & \multicolumn{1}{c|}{ 0.891 } 
 &  0.880  
 & \multicolumn{1}{c|}{ \textbf{0.688} } 
 & \multicolumn{1}{c|}{ 0.982 } 
 &  0.973  
 & \multicolumn{1}{c|}{ 10.3 } 
 &  3.8 \\\hline
\multirow{3}{*}{Hepatic Vessels}  
 &  nnInteractive  
 & \multicolumn{1}{c|}{ 0.113 } 
 & \multicolumn{1}{c|}{ 0.165 } 
 &  0.209  
 & \multicolumn{1}{c|}{ 0.122 } 
 & \multicolumn{1}{c|}{ 0.232 } 
 &  0.289  
 & \multicolumn{1}{c|}{ 82 } 
 &  79.8 \\\cline{2-10}
  
 &  CLoPA-I.N  
 & \multicolumn{1}{c|}{ \textbf{0.407} } 
 & \multicolumn{1}{c|}{ 0.555 } 
 &  0.556  
 & \multicolumn{1}{c|}{ 0.542 } 
 & \multicolumn{1}{c|}{ \textbf{0.716} } 
 &  \textbf{0.715}  
 & \multicolumn{1}{c|}{ 39.4 } 
 &  31.7 \\\cline{2-10}
  
 &  CLoPA-C.N  
 & \multicolumn{1}{c|}{ 0.405 } 
 & \multicolumn{1}{c|}{ \textbf{0.562} } 
 &  \textbf{0.563}  
 & \multicolumn{1}{c|}{ 0.542 } 
 & \multicolumn{1}{c|}{ 0.709 } 
 &  0.710  
 & \multicolumn{1}{c|}{ \textbf{37.5} } 
 &  \textbf{30.3} \\\hline
\multirow{3}{*}{Colon Cancer}  
 &  nnInteractive  
 & \multicolumn{1}{c|}{ 0.475 } 
 & \multicolumn{1}{c|}{ 0.728 } 
 &  0.725  
 & \multicolumn{1}{c|}{ 0.520 } 
 & \multicolumn{1}{c|}{ 0.788 } 
 &  0.790  
 & \multicolumn{1}{c|}{ 2.4 } 
 &  0 \\\cline{2-10}
  
 &  CLoPA-I.N  
 & \multicolumn{1}{c|}{ \textbf{0.576} } 
 & \multicolumn{1}{c|}{ 0.783 } 
 &  0.779  
 & \multicolumn{1}{c|}{ \textbf{0.660} } 
 & \multicolumn{1}{c|}{ 0.867 } 
 &  0.867  
 & \multicolumn{1}{c|}{ \textbf{1.7} } 
 &  0 \\\cline{2-10}
  
 &  CLoPA-C.N  
 & \multicolumn{1}{c|}{ 0.572 } 
 & \multicolumn{1}{c|}{ \textbf{0.784} } 
 &  0.778  
 & \multicolumn{1}{c|}{ 0.658 } 
 & \multicolumn{1}{c|}{ \textbf{0.872} } 
 &  \textbf{0.869}  
 & \multicolumn{1}{c|}{ 2 } 
 &  0.256 \\\hline
\end{tabular}
}
\label{tab:experiment_results_traj}
\end{table}

\noindent \textbf{Trajectory analysis:} Table~\ref{tab:experiment_results_traj} confirms the same trends across the full dataset-size trajectory, though improvements are less pronounced than in the final-episode snapshot because the trajectory AUC averages over all episodes, including before adaptation takes effect. As shown in Fig.~\ref{fig:final_dice_trajectory_selection}, at least one adaptive method is always capable of reaching nnU-Net performance on unattainable tasks for the base model (hepatic vessel and hippocampus). Moreover, we see that instance normalisation tuning rapidly elevates expected performance to near expert-level after the first training episode, but subsequently plateaus across all tasks. This is practically significant: it means that clinicians benefit from improved predictions early in the annotation campaign, reducing cumulative user effort over the full dataset. Since the initial episode yields the majority of the benefit, a two-phase strategy---triggering instance normalisation adaptation early, then transitioning to deeper feature-representation tuning as more data becomes available---would likely yield further gains. This is supported by the adaptation of convolution kernels producing higher peak performances, but being less stable in small data regimes. Such a curriculum could also incorporate dynamic prompting strategies, for instance extending the interaction window, to maximise the quality of the training signal provided to the model. In summary, low-parameter adaptation elevates performance ceilings, stabilises editing behaviour, boosts annotation efficiency and enables specialist performance to be reached on all tasks using low-effort clicking, with only a fraction of the data.

